\documentclass{article}

\usepackage{microtype}
\usepackage{graphicx}
\usepackage{subfigure}
\usepackage{booktabs} 

\usepackage{hyperref}

\usepackage[accepted]{arxiv}

\usepackage{amsmath}
\usepackage{amssymb}
\usepackage{mathtools}
\usepackage{amsthm}

\usepackage[capitalize,noabbrev]{cleveref}

\theoremstyle{plain}

\theoremstyle{definition}

\theoremstyle{remark}

\usepackage[textsize=tiny]{todonotes}

\icmltitlerunning{MARLIN: Mixed-Precision Auto-Regressive Parallel Inference on Large Language Models}

\usepackage[ruled,linesnumbered]{algorithm2e} %
\usepackage{algorithmic}
\usepackage{listings}

\usepackage{subcaption}

\usepackage{multirow}

\usepackage{tikz}
\newcommand\encircles[1]{%
  \tikz[baseline=(X.base)]
    \node (X) [font=\fontsize{8}{0}\selectfont,draw, shape=circle, inner sep=0, fill=black, text=white] {\strut #1};%
}

\newcommand{\code}[1]{\texttt{\small{#1}}}

\newcommand{\myvec}[1]{\pmb{#1}}

\newcommand{\zeros}{z}
\newcommand{\scale}{s}
\newcommand{\bits}{b}

\renewcommand{\paragraph}[1]{\vspace{0.1em}\noindent\textbf{#1}}

\begin{document}

\twocolumn[
\icmltitle{MARLIN: Mixed-Precision Auto-Regressive \\
Parallel Inference on Large Language Models}



\icmlsetsymbol{equal}{*}

\begin{icmlauthorlist}
\icmlauthor{Elias Frantar}{ista}
\icmlauthor{Roberto L. Castro}{udc}
\icmlauthor{Jiale Chen}{ista}
\icmlauthor{Torsten Hoefler}{ethz}
\icmlauthor{Dan Alistarh}{ista,neuralmagic}
\end{icmlauthorlist}

\icmlaffiliation{ista}{Institute of Science and Technology Austria (ISTA), Klosterneuburg, Austria}
\icmlaffiliation{udc}{CITIC, Universidade da Coruña, A Coruña, Spain}
\icmlaffiliation{ethz}{D-INFK, ETH Zurich, Zurich, Switzerland}
\icmlaffiliation{neuralmagic}{Neural Magic, Inc., Somerville, United States}

\icmlcorrespondingauthor{Jiale Chen}{jiale.chen@ist.ac.at}

\icmlkeywords{Large language model (LLM) inference, GPU programming, Batch parallelism}

\vskip 0.3in
]



\printAffiliationsAndNotice{}  

\begin{abstract}
As inference on Large Language Models (LLMs) emerges as an important workload in machine learning applications,  weight quantization has become a standard technique for efficient GPU deployment. Quantization not only reduces model size, but has also been shown to yield substantial speedups for single-user inference, due to reduced memory movement, with low accuracy impact. Yet, it remains open whether speedups are achievable also in \emph{batched} settings with multiple parallel clients, which are highly relevant for practical serving. It is unclear whether GPU kernels can be designed to remain practically memory-bound, while supporting the substantially increased compute requirements of batched workloads.

This paper resolves this question positively by describing the design of Mixed-precision Auto-Regressive LINear kernels, called MARLIN. Concretely, given a model whose weights are compressed via quantization to, e.g., 4 bits per element, MARLIN shows that batchsizes up to 16-32 can be supported with close to maximum ($4\times$) quantization speedup, and larger batchsizes up to 64-128 with gradually decreasing, but still significant, acceleration. MARLIN accomplishes this via a combination of techniques, such as asynchronous memory access, complex task scheduling and pipelining, and bespoke quantization support. Our experiments show that MARLIN's near-optimal performance on individual LLM layers across different scenarios can also lead to end-to-end LLM inference speedups (of up to $2.8\times$) when integrated with the popular vLLM serving engine. Finally, MARLIN is extensible to further compression techniques, like NVIDIA 2:4 sparsity, leading to additional speedups.
\end{abstract}

\section{Introduction}

\sloppy
The capabilities of large language models (LLMs)~\cite{radford2019language, zhang2022opt, touvron2023llama} have led to significant research and industrial interest. Consequently, a lot of effort has been dedicated to reducing their computational costs, and notably their inference costs \cite{dettmers2022llm, frantar2022gptq, lin2023awq, xiao2022smoothquant, dettmers2022case, shao2023omniquant, sheng2023flexgen}. 
A large fraction of this work starts from the observation that \emph{generative} workloads---in which a model produces a next token (often a word) based on a cached context---can be heavily memory-bound when executed on GPUs or CPUs, as the cost of reading the LLM weights dwarfs that of the arithmetic operations, and their footprint greatly exceeds the cache size.  

Reducing memory movement leads to substantial practical speedups by \emph{compressing the network weights}, as shown by various recent works \cite{frantar2022gptq, lin2023awq, shao2023omniquant, dettmers2023spqr}, in particular in the context of quantization. Specifically, during inference, weights can often be loaded from GPU memory in compressed form---reducing movement costs---and then dynamically decompressed in registers before multiplication.


A key limitation of existing such \emph{mixed-precision} inference implementations is that they cease to provide significant speedups in the \emph{batched inference} case, that is, when multiple tokens must be generated in parallel.
Intuitively, this is because this case has significantly higher arithmetic intensity, making it much harder to fully hide all required computations behind the reduced memory movement.

Yet, the batched scenario is key in large-scale LLM applications: for instance, OpenAI is claimed to produce 100 billion words a day~\cite{SamA}--that is, more than 1 million words a second--providing ample opportunities for parallelism, and in fact \emph{the necessity} for grouping these requests to achieve highest GPU utilization. 

\paragraph{Contribution.} 
In this work, we investigate software support for LLM inference acceleration via mixed-precision in the general \emph{batched} case. 
We observe that, across GPU types, quantized LLM generative inference \emph{remains memory-bound} even for fairly large input sizes: in practice, one could still obtain close to the full speedup from reduced memory movement even when 16-32 tokens are generated in parallel.

Concretely, this is because modern GPUs, such as the ones from NVIDIA's Ampere family, typically have a FLOP-to-byte ratio in the range of 100 to 200 for FP16 operations~\cite{ampere}. 
Thus, if one would be able to reduce weight precision to 4 bits while maintaining a proportional number of multiply-accumulate operations per quantized weight (in this case, in the range of 25-50), one could theoretically still obtain close to the optimal $4\times$ speedup. 
Yet, realizing this in practice is complex.

In this paper, we present the design and implementation of a family of mixed-precision inference kernels called MARLIN, which achieve near-optimal batched inference speedups due to reduced memory movement on modern, widely available, NVIDIA Ampere GPUs. MARLIN kernels combine various techniques, ranging from advanced task scheduling, partitioning, and pipeplining techniques to quantization-specific layout and compute optimizations.

\begin{figure}[t]
    \centering
    \includegraphics[width=0.9\linewidth]{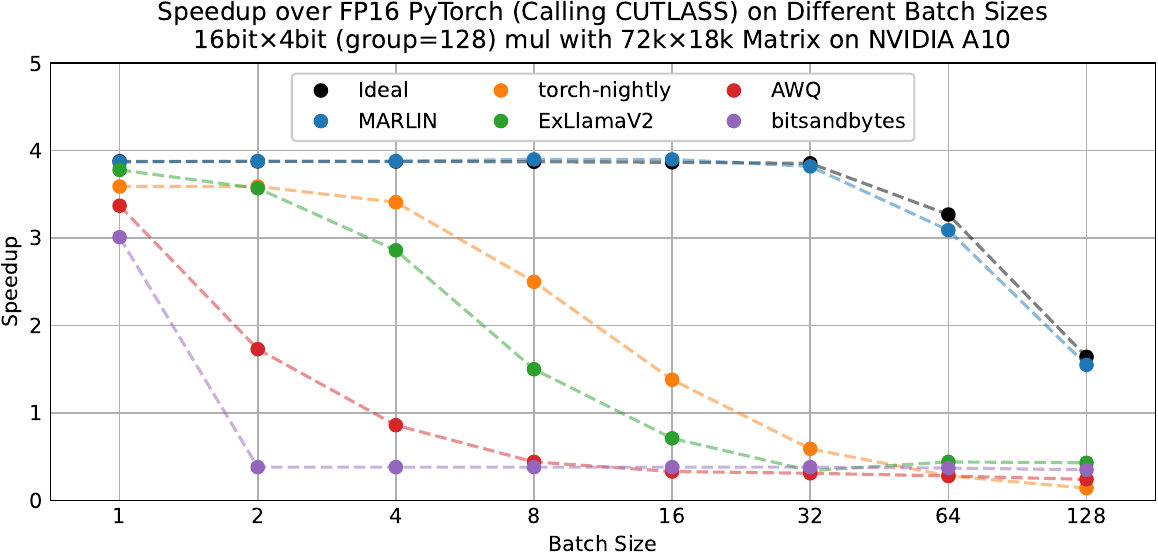}
    \caption{Illustration of MARLIN peak performance while increasing batch size, for a single large linear LLM layer, compared with other popular open-source kernels, showing that we can achieve near-optimal performance in this scenario.}
    \label{fig:marlin-peak}
\end{figure}

We validate our design both via individual per-layer benchmarks, and end-to-end through an integration with vLLM~\cite{vllm}, a popular open-source LLM serving engine. 
Specifically, for 4bit-weight inference, MARLIN obtains speedups of approximately $3.9\times$ relative to FP16 on an inference-optimized NVIDIA A10 GPU and large matrices, for batch sizes of up to 16-32. (See Figure~\ref{fig:marlin-peak}). 
Speedups gradually reduce, towards $1.5\times$ at batch size 128, as the problem becomes compute-bound. 
Our analysis shows that this is close to optimal. 
In addition to the base design, 
we present Sparse-MARLIN, an extension of MARLIN to the 2:4-sparse Tensor Core format, which provides additional speedups of up to {65\%} relative to the original (dense) variant.

We also extend our benchmarks to end-to-end (full model) results in an industrial inference setting, via a vLLM~\cite{vllm} integration, on top of leading open LLMs such as Llama~\cite{touvron2023llama2} and Falcon~\cite{falcon2023}, for which accurate 4bit quantization and 2:4 sparsification is possible.
According to our end-to-end measurements, the MARLIN kernel dramatically increases the speed of multi-user token generation, achieving up to a $2.8\times$ speedup compared to vLLM’s standard precision kernel, at batchsize 16. Sparse-MARLIN further improves performance, providing speedups of up to $3.2\times$.


Overall, we show that near optimal weight-quantized LLM inference speedups can be achieved also at batchsizes significantly larger than 1. This is done via a new kind of GPU kernel design, which takes full advantage of hardware capabilities specifically mixed-precision, and should be extensible to other compression formats. The code for MARLIN\footnote{https://github.com/IST-DASLab/marlin} and its Sparse-MARLIN variant~\footnote{https://github.com/IST-DASLab/Sparse-Marlin} are available openly, as well as the vLLM integration~\footnote{https://github.com/vllm-project/vllm}.  


\section{Background}

We continue with an overview of GPU architecture, and the CUDA programming and execution model. We  focus on the Tensor Core improvements introduced by NVIDIA Ampere, which we utilize extensively. Finally, we provide some background on mixed-precision inference in LLMs.

\subsection{Graphics Processing Units}
NVIDIA GPUs comprise an array of Streaming Multiprocessor (SM) elements that share a DRAM memory, known as Global MEMory (GMEM) and an L2 cache.
Each SM is divided into partitions, which contain various processing blocks. Each processing block includes a warp scheduler, a Register File (RF), and an L0 instruction cache.
The processing blocks within an SM share an L1 cache, which can be partially reconfigured as a fast scratch pad memory referred to as Shared Memory (SMEM).
Within each processing block, there are four types of units: Integer Units, Special Function Units, Floating-Point Units (FPU) / CUDA Cores, and Tensor-Core Units (TCU).

TCUs, first introduced in the Volta architecture, primarily target ML workloads by enabling one matrix multiply-and-accumulate (MMA) operation per cycle. This reduces the cost of fetching and decoding multiple instructions needed for such computations. In the Ampere architecture, TCUs can deliver up to $16\times$ more performance on FP16 than fused multiply-add (FMA) operations running on FPUs.

The CUDA programming and execution model is closely related to the architecture specifics described. 
It defines three granularity levels, encompassing thread blocks, warps, and threads.
The warp is the basic scheduling unit in CUDA, consisting of $32$ threads that are executed concurrently.
Thread blocks are a collection of warps, scheduled for execution on the same SM. The number of warps, and the number of thread-blocks running simultaneously on each SM is contingent upon hardware limitations, such as the number of warp schedulers, registers per thread, or the SMEM available. 

\subsubsection{Modern Tensor Core Units}

\begin{figure}[ht]
    \centerline{\includegraphics[width=0.85\linewidth]{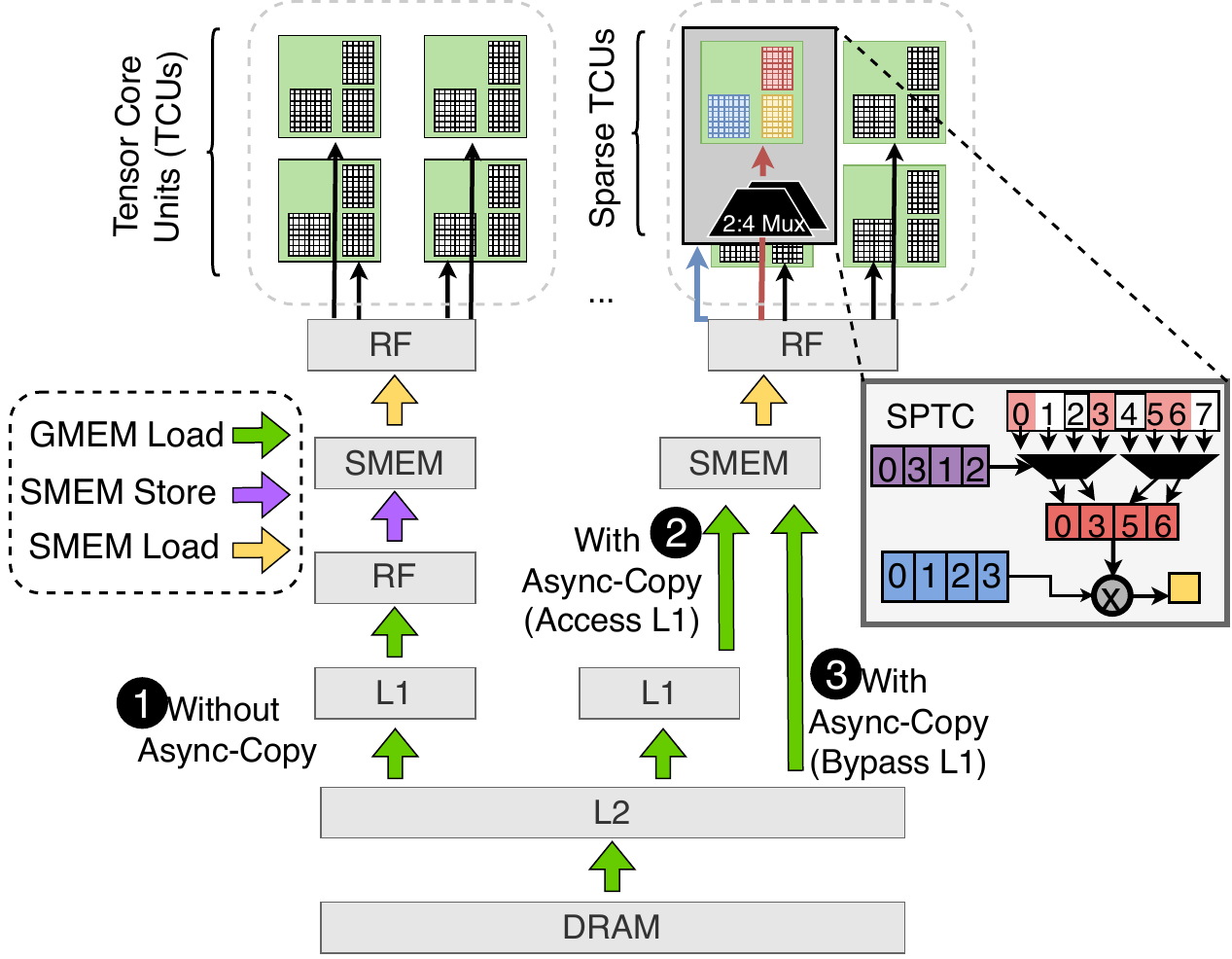}}
    \caption{Illustration of asynchronous copy operation with and without L1 bypass (right) vs. standard operations (left).}
    \label{fig:async}
\end{figure}

Ampere GPUs extended their TCUs with respect to previous generations to handle both 1) fine-grained structured sparsity, resulting in Sparse Tensor-Core Units (SPTCs), and 2) asynchronous copy operations. First, structured sparsity is supported through a new 2:4 format, promising a $2\times$ speedup over the original TCUs, and up to $32\times$ over FPUs.

The 2:4 format divides the LHS matrix into vectors of length four, and for each vector it zeros-out two elements, resulting in a $50\%$ sparse but structured matrix. Figure~\ref{fig:async} shows a simplified representation of an SPTC. 
Two data structures will represent the sparsified matrix: (1) a values structure, depicted in blue, containing the non-zero values, (2) a metadata structure, depicted in purple, containing the position of each non-zero value within each group of $4$ elements. 
The metadata structure will be used by the new hardware components on SPTCs to select just the elements of the RHS matrix that are needed in the computation, skipping the zeroed-out values.

NVIDIA's Ampere microarchitecture introduces data fetching improvements for enhanced Tensor Core performance.  This involves a new asynchronous copy instruction that allows loading data directly from GMEM into SMEM. 
As shown in Figure~\ref{fig:async} \encircles{1}, in previous generations it was necessary to first load through L1 cache into RF with global load instructions. 
Then, the data was transferred to SMEM with shared store instructions, and finally moved into RF with shared load instructions.

Ampere's new asynchronous copy saves SM internal bandwidth by avoiding intermediate RF access. 
There are two variants of this instruction, ``access'' that saves data into L1 for subsequent accesses and reuse (Figure~\ref{fig:async} \encircles{2}), and ``bypass'', which also skips L1 cache (Figure~\ref{fig:async} \encircles{3}).


\subsection{Mixed-Precision Inference on LLMs}

Mixed-precision LLM inference offers the potential to reduce a model's large memory footprint, and correspondingly accelerate memory-bound workloads by statically compressing pretrained model weights while decompressing them on-the-fly during inference as needed.



\paragraph{Weight Quantization}. 
A standard LLM compression approach is weight-only quantization, which reduces the precision in which the weights $\boldsymbol{W}$ are stored, while leaving the layer inputs $\boldsymbol{X}$ untouched. 
This is extremely popular, e.g.,~\cite{frantar2022gptq, lin2023awq, dettmers2022case, dettmers2022llm, dettmers2023spqr},
 as it has shown remarkable accuracy robustness even at relatively high compression rates.

Broadly, weight quantization lossily compresses floating-point weights by mapping them to a limited set of integer levels. We focus on uniform quantization, meaning that given a vector $\myvec{v} \in \mathbb{R}^n$, we define 
\begin{align}
\label{eq:quantization}
Q \left( \myvec{v}, b \right)
&= \left\lfloor \frac{\myvec{v} - \min\left(\myvec{v}\right)}{\max\left(\myvec{v}\right) - \min\left(\myvec{v}\right)} \left( 2^\bits - 1 \right) \right\rceil \nonumber = \left\lfloor \left( \myvec{v} - \zeros \right) / \scale \right\rceil,
\end{align}
where $\lfloor \cdot \rceil$ rounds to nearest, $\zeros = \zeros \left(\myvec{v}\right) = \min\left(\myvec{v}\right)$ maps to zero and $\scale = \scale\left(\myvec{v}\right)
= \left( \max\left(\myvec{v}\right) - \min\left(\myvec{v}\right) \right) / \left( 2^\bits -1 \right)$ is the scale. 

The reconstruction error can be computed as $\varepsilon_r = \left\| \myvec{v} - Q \left( \myvec{v}, b \right) \right\|_2$. 
We can improve the error by partitioning $\myvec{v}$ into groups and quantizing each group separately, thus storing $\scale$ and $\zeros$ values for each group, e.g., of 128 contiguous values.  

In this paper, we use perform the actual weight quantization via a variant of GPTQ~\cite{frantar2022gptq}, which takes advantage of second-order information to compensate for quantization errors, allowing for only minor accuracy degradation. However, we emphasize that our kernel techniques are independent of any particular quantization algorithm.


\section{The MARLIN Kernel} 
\label{sec:marlin-kernel}

\subsection{Motivation}

LLM weight quantization is motivated by the fact that modern GPUs have large FLOPs/Bytes ratios, meaning that they can execute floating point operations much faster than they can read from memory. As an example, an A10 GPU has a FLOPs/Bytes ratio of $\approx 200$ \cite{nvidia-a10}.
In the context of a single layer matrix multiplication, processing one input token takes 2 FLOPs per weight and the GPU can execute 100 FLOPs in the time it takes to load one 4-bit weight.
Hence, memory loading will dominate runtime as long as the input batchsize is less than $b_\text{opt} \approx 50$. In fact, $b_\text{opt}$ is the batchsize where latency is neither bound by memory nor by compute, i.e., where we achieve the \emph{lowest latency at maximum throughput}. In principle, this is precisely the batchsize that we would like to operate at in practice: any smaller does not yield further speedup and any larger does not improve throughput. 
(This analysis is further detailed in Section~\ref{sec:roofline}.) 

However, actually implementing such a mixed-precision (FP16-INT4) matrix multiplication (matmul) kernel which fully maximizes essentially all GPU resources, i.e., compute and memory, \emph{simultaneously}, is a major challenge. In the following, we will try to come as close as possible to this goal by designing \emph{MARLIN}, an extremely optimized \emph{Mixed-precision Auto-Regressive LINear} kernel.

\subsection{Ampere Matrix Multiplication}

We begin by describing the general concepts used to implement peak performing (uniform precision) matrix multiplication kernels on GPUs, in particular on Ampere class devices. We closely follow the CUTLASS hierarchical parallelization model \cite{cutlass-gemm}. Concretely, we consider the problem of multiplying an $M \times K$ matrix $\mathbf{A}$ with a $K \times N$ matrix $\mathbf{B}$ to produce an $M \times N$ output matrix $\mathbf{C}$.

\paragraph{SM Level.} As a first step, $\mathbf{A}$ is partitioned into $M_\text{sm} \times K_\text{sm}$ blocks $\mathbf{A}_\text{sm}[i_\text{sm}, k_\text{sm}]$, $\mathbf{B}$ into $K_\text{sm} \times N_\text{sm}$ blocks $\mathbf{B}_\text{sm}[k_\text{sm}, j_\text{sm}]$ and $\mathbf{C}$ into $M_\text{sm} \times N_\text{sm}$ blocks $\mathbf{C}_\text{sm}[i_\text{sm}, j_\text{sm}]$. Due to the nature of a matrix multiplication, all $\mathbf{C}_\text{sm}[i_\text{sm}, j_\text{sm}]$ can be computed independently by accumulating the results of $\mathbf{A}_\text{sm}[i_\text{sm}, k_\text{sm}] \mathbf{B}_\text{sm}[k_\text{sm}, j_\text{sm}]$ over all $k_\text{sm}$. Consequently, computation can be easily parallelized by distributing those $\mathbf{C}_\text{sm}[i_\text{sm}, j_\text{sm}]$ sub-problems across the GPU's independent compute units, its SMs. At this stage, $\mathbf{A}_\text{sm}[i_\text{sm}, k_\text{sm}]$ and $\mathbf{B}_\text{sm}[k_\text{sm}, j_\text{sm}]$ blocks must be loaded from global GPU memory. Similarly, $\mathbf{C}_\text{sm}[i_\text{sm}, j_\text{sm}]$ must eventually be written back to global storage, but intermediate accumulation can happen directly in registers, as we will discuss next.

\paragraph{Warp Level.} Within the sub-problem considered by a single SM, another equivalent partitioning, this time with parameters $M_\text{wa}$, $K_\text{wa}$, and $N_\text{wa}$, is performed. This is in order to assign independent $\mathbf{C}_\text{wa}[i_\text{sm}, j_\text{sm}][i_\text{wa}, j_\text{wa}]$ output accumulation tasks to different warps. Crucially, the SM blocks $\mathbf{A}_\text{sm}[i_\text{sm}, k_\text{sm}]$ and $\mathbf{B}_\text{sm}[k_\text{sm}, j_\text{sm}]$ can be temporarily stored in shared memory, so that the repeated loading of $\mathbf{A}_\text{wa}[i_\text{sm}, k_\text{sm}][i_\text{wa}, k_\text{wa}]$ and $\mathbf{B}_\text{wa}[k_\text{sm}, j_\text{sm}][k_\text{wa}, j_\text{wa}]$ by different warps is much faster. Meanwhile, outputs $\mathbf{C}_\text{wa}[i_\text{sm}, j_\text{sm}][i_\text{wa}, j_\text{wa}]$ are kept in the corresponding warp's registers, eliminating any additional memory access costs during accumulation.

\paragraph{Tensor Core Level.} Eventually, each warp will repeatedly multiply $M_\text{wa} \times K_\text{wa}$ and $K_\text{wa} \times N_\text{wa}$ matrices. While the corresponding matrix dimensions are small at this level, they still typically exceed the fundamental Tensor Core $(M_\text{tc}, K_\text{tc}, N_\text{tc})$ shape. Consequently, one final partitioning step is required. However, unlike before, $\mathbf{C}_\text{tc}[i_\text{sm}, j_\text{sm}][i_\text{wa}, j_\text{wa}][i_\text{tc}, j_\text{tc}]$ are accumulated \emph{sequentially} by a single warp. While all data is in registers at this point and there is thus no memory access cost, it is still important to perform the loop over $k_\text{tc}$ \emph{outermost}. This is to remove sequential dependencies between Tensor Core operations as much as possible to maximize throughput. It should be noted that actually utilizing Tensor Cores requires further distribution of matrix elements across threads in very specific patterns. However, this is a technical detail mandated by the microarchitecture rather than another flexible opportunity for parallelization.

\subsection{Mixed-Precision Challenges}

Adapting the above uniform precision matmul to the mixed-precision case while maintaining peak performance, in particular for medium $M$ where the operation is (close to) memory-bound, is challenging for the following reasons:

\begin{enumerate}
    \item The various parallelization levels must be very carefully configured to ensure that the loading of the quantized operand $\mathbf{B}$ actually is the kernel's main runtime bottleneck; and not, e.g., repeated reloading of full precision $\mathbf{A}_\text{sm}$ blocks.
    \item As runtime is dominated by memory loading, this aspect must hit peak efficiency, despite the significantly compressed representation of $\mathbf{B}$.
    \item For medium $M$, the cost of matmul computations can get close to the overall the memory loading cost, hence requiring extremely careful overlapping to stay close to theoretical performance. Additionally, we also need to manage quantization metadata, making this part even more tricky.
    \item Partitioning constraints forced by Challenge 1, together with the fact that $M$ is not very large, significantly limit parallelization options. This makes it hard to achieve peak memory loading and compute on both the SM \emph{and} warp level, respectively. This effect is amplified further by existing model matrix shapes which can be unfavorable for specific GPUs.
\end{enumerate}

Our MARLIN kernel specifically addresses all of the above challenges, eventually allowing it to achieve close to peak performance in many practical settings.




\subsection{Kernel Design}

In what follows, we assume that the matrix $\mathbf{A}$ is in full FP16 precision, while the $K\times N$ matrix $\mathbf{B}$ has been (symmetrically) quantized to INT4, either with one FP16 scale for each of the $N$ columns or one scale per $G$ consecutive weights in each column, for $\lceil K/G \rceil N$ scales in total.

\paragraph{Bound By Weight Loading.}
Executing our target matmul requires, in theory, touching exactly $16MK + 4KN + 16MN$ bits of memory (reading both operands and writing the results) while executing exactly $MKN$ multiply-accumulate operations, each counted as 2 FLOPs. If $M$ is relatively small, our problem has low arithmetic intensity. Consequently, it should be bound by the cost of reading the quantized weights $\mathbf{B}$ from global GPU memory.

This holds in theory, but we need to organize computation carefully for this to remain true in practice. In contrast to the previously studied \cite{frantar2022gptq, dettmers2022case} $M = 1$ case, where both $\mathbf{A}$ and $\mathbf{C}$ are tiny, inputs and outputs now actually have non-negligible size, especially since those operands have $4\times$ higher bit-width than our weights. Hence, we need to pick a sufficiently large $N_\text{SM}$ to minimize costly reloading of $\mathbf{A}_\text{sm}$ blocks. At the same time, this reduces the number of $\mathbf{C}_\text{sm}[i_\text{sm}, j_\text{sm}]$ sub-problems, making it hard to fully utilize all SMs.

The key to working around these problems is exploiting the GPU's \emph{L2 cache}, which is usually significantly faster than global memory. Additionally, a GPU can load from L2 to L1 and from global to L2 simultaneously. Thus, we can pipeline these loads and essentially hide the bandwidth cost of the $\mathbf{A}_\text{sm}$ block loads completely, as long as the overall required memory traffic does not exceed the L2 bandwidth. Consequently, we will proceed by partitioning $\mathbf{C}$ into blocks of size $M \times N_\text{sm}$ with $N_\text{sm} \in \{64, 128, 256\}$, i.e., moderately wide tiles of full input batchsize, and then assigning each corresponding independent matmul sub-problem to one SM. At $N_\text{sm} = 256$, even batchsize $M = 64$ remains bound by global weight loading. More precisely, global loading of $\mathbf{A}_\text{sm}$ blocks remains the bottleneck as long as reading both $\mathbf{A}_\text{sm}$ and $\mathbf{B}_\text{sm}$ blocks from L2 is faster, i.e.:
\begin{equation}
    (2 M K_\text{sm} + 0.5 K_\text{sm} N_\text{sm}) / B_\text{l2} < (0.5 K_\text{sm} N_\text{sm}) / B_\text{gl},
\end{equation}
where $B_\text{l2}$ and $B_\text{gl}$ denote the L2 and global bandwidth, respectively.

\paragraph{Maximizing Loading Bandwidth.}
In order to maximize practical loading bandwidth, we aim to utilize the widest loads possible; on current GPUs 16 bytes (128 bits) per thread. This means one warp can load $32 \times 32 = 1024$ INT4 weights with a single instruction. To reach peak efficiency, we need to have 8 threads each in a warp read 128 bytes of contiguous chunks from GMEM (assuming 128-byte-aligned addresses), a full cache line. Achieving this for $\mathbf{A}$ blocks of shape $M \times K_\text{sm}$ mandates a $K_\text{sm}$ of at least 64. Since the weights are static during inference and can thus be preprocessed offline, we simplify things by reshuffling $16 \times 64$ tiles so that they are laid out contiguously in memory and are thus loaded optimally, which also simplifies corresponding indexing.

While we continuously reload $\mathbf{A}_\text{sm}$ blocks from L2 cache, each element of $\mathbf{B}$ is accessed exactly once. Nevertheless, every read will always be put into the L2 cache, potentially evicting parts of $\mathbf{A}$ that are still needed by some SMs. To avoid such cache pollution, we use the \code{cp.async} instruction with an \code{evict\_first} cache-hint, ensuring that unnecessarily stored $\mathbf{B}$ data is dropped before any other cache line.

\paragraph{Shared Memory Layouts.}
Overall, we always load asynchronously via Ampere's \code{cp.async} instruction from global (or L2) to shared memory; this requires no temporary registers and also makes overlapping these loads with computation much easier. Due to our offline preprocessing of $\mathbf{B}$, we can simply copy to shared memory in contiguous fashion, avoiding bank conflicts.

In contrast, handling the $\mathbf{A}$ fragments requires a lot more care: specifically, we need to ensure that the 16-byte vectors corresponding to indices $ij$, $(i + 8)j$, $i(j + 1)$ and $(i + 8)(j + 1)$ of each $16\times 16$ FP16 $\mathbf{A}$ block are stored in different memory banks. Only then can \code{ldmatrix.sync} instructions execute in conflict-free manner. (Those load $\mathbf{A}$ operand data and distribute it across warp threads to prepare for Tensor Core use.)
This can be achieved by storing 16-byte element $ij$ in an activation tile at location $i(i \oplus j)$ in the corresponding shared memory tile, where $\oplus$ denotes the XOR operation \cite{cutlass-xor}. Another key aspect of this index transformation is that if a warp reads a contiguous sub-tile of the global $\mathbf{A}$ tile (e.g., the first 4 rows), then it will be written permuted but still overall contiguously into shared memory. Although undocumented, this appears to be necessary in order to avoid bank conflicts on writing, as we observed when analyzing outputs of the NVIDIA profiler.

These index calculations are somewhat complex and potentially slow to take care of dynamically; however, as they only affect a relatively small number of shared memory locations, which remain static throughout the main loop, we can precompute them in registers, accompanied by appropriate unrolling, described below.

\begin{figure}[th]
    \centering
    \includegraphics[width=.3\textwidth]{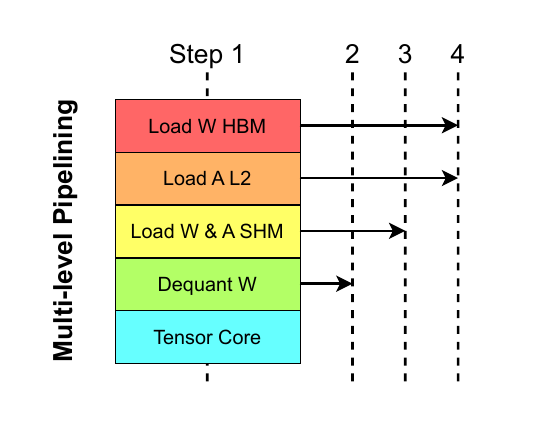}
    \caption{Levels of pipelining in the MARLIN kernel.}
    \label{fig:marlin-pipelining}
\end{figure}

\paragraph{Memory Load Pipelining.}
The key to simultaneously reaching close to maximum bandwidth and close to maximum compute is to fully overlap memory loading and Tensor Core math. For global to shared memory loads, this can be achieved via \code{cp.async} operations, in every iteration prefetching the $\mathbf{A}_\text{sm}$ and $\mathbf{B}_\text{sm}$ blocks which will be used $P - 1$ steps in the future, where $P$ is the pipeline depth (we need one more buffer for the current tile). Additionally, we can prefetch the next sub-tile from shared memory (most GPU operations do not block until they hit a dependency) before accumulating the current partial matmul, for which the operands were already fetched to registers in the previous iteration---this technique is also called \emph{double buffering} \cite{cutlass-gemm}. We pick a pipeline depth of $P = 4$ for two reasons: (a) this seemed sufficient in all of our tests to completely hide latency while fitting into shared memory even for $M = 64$, and (b) because it is evenly divisible by 2. The latter is crucial as it allows us to smoothly unroll across the full pipeline since after $P$ iterations both the pipeline and the register buffer index will always have the same value of 0. This unrolling makes all shared memory addressing completely static, avoiding slow transformed index calculations (see above) by using some of the extra registers that we have available. Finally, we would like to note that this also seemed to be the most reliable way to make the CUDA compiler correctly order instructions to enable actual double buffering. Figure~\ref{fig:marlin-pipelining} visualizes the several layers of pipelining used by the MARLIN kernel.

\begin{figure}[h!]
    \centering
    \includegraphics[width=.33\textwidth]{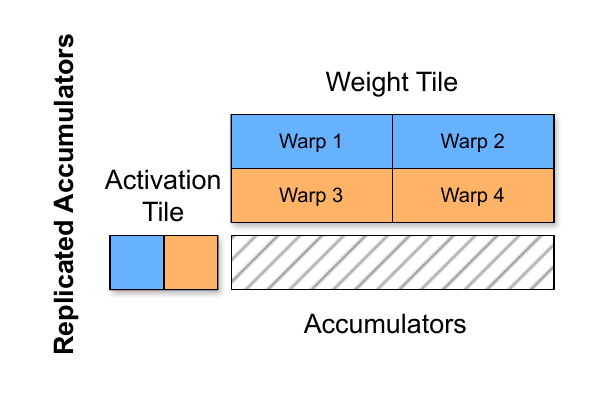}
    \caption{Illustration of MARLIN's warp layout. Multiple warps accumulate partial results of the same output tile; see also Algorithm~\ref{alg:warp-layout} for corresponding pseudocode.} 
    \label{fig:marlin-warps}
\end{figure}

\paragraph{Warp Layout.}
The computation of $\textbf{C}_\text{sm}$ on a single SM must further be subdivided across warps: if done in direct fashion, each warp would compute an $M \times (N_\text{sm} / \text{\#warps})$ tile of the output. In order to reach peak compute throughput, we would like to use at least four (as Ampere GPUs have four warp schedulers) and ideally eight warps \cite{sun2022dissecting}, to have additional latency hiding. However, this leads to small tile sizes, especially at smaller $N_\text{sm}$. This is not only problematic for our memory reshuffling discussed above but also hinders Tensor Core throughput since a small tile-width brings more sequential dependencies (as those consecutive operations must use the same accumulators) into tensor-core operations, which can cause stalls. Instead, we fix the sub-tile width of each warp to 64 and further split the computation across $K_\text{sm}$; Figure~\ref{fig:marlin-warps} illustrates such a warp layout and Algorithm~\ref{alg:warp-layout} provides corresponding pseudo-code. Consequently, multiple warps will accumulate partial results of the same $\mathbf{C}_\text{wa}[i_\text{sm}, j_\text{sm}][i_\text{wa}, j_\text{wa}]$ in registers. These must then eventually be reduced in shared memory before the final write-out. Yet, this can be done via a logarithmic parallel reduction \cite{harris2007optimizing}, which typically causes minimal overhead.

\begin{algorithm}
    \centering
    \caption{Warp reduction within a corresponding $\mathbf{C}_\text{sm}[i_\text{sm}, j_\text{sm}]$ sub-problem. The following pseudo-code is executed by all warps, identified via ``warp\_idx''.}
    \begin{algorithmic}
        \STATE $\mathbf{C} \gets$ all zeros matrix of shape $M_\text{wa} \times N_\text{wa}$
        \STATE $j \gets \text{warp\_idx} \,\, \text{mod} \,\, (N_\text{sm} / N_\text{wa})$
        \FOR{$k_\text{sm} \gets 1, \dots, K / K_\text{sm}$}
            \STATE $i \gets \lfloor \text{warp\_idx} / (M_\text{sm} / M_\text{wa}) \rfloor$
            \WHILE{$i < K_\text{sm} / K_\text{wa}$}
                \STATE $\mathbf{C} \gets \mathbf{C} + \mathbf{A}_\text{wa}[i_\text{sm}, j_\text{sm}][0, i] \mathbf{B}_\text{wa}[i_\text{sm}, j_\text{sm}][i, j]$
                \STATE $i \gets i + \text{\#warps} / (N_\text{sm} / N_\text{wa})$
            \ENDWHILE
        \ENDFOR
        \STATE $\mathbf{C}_\text{wa}[i_\text{sm}, j_\text{sm}][0, j] \gets$ parallel reduction of $\mathbf{C}$ across $j$
    \end{algorithmic}
    \label{alg:warp-layout}
\end{algorithm}

\paragraph{Dequantization and Tensor Cores.}
Doing naive type-casts from INT4 to FP16 is slow; instead, we follow a modified version of the binary manipulations of~\citet{kim2022says}.

We now illustrate this procedure in the simplest case: converting the INT4 located at positions $12 - 15$ in an INT16 to a signed FP16 value. First, we extract just the bits corresponding to our INT4 (via an AND of a mask) and turn bits $1-7$ of the result into 0110010 (with an OR); this can be accomplished in a single \lstinline|lop3| instruction, which we however seemingly need to emit explicitly. Now, we have an FP16 number with an exponent of 50 and the last 4 mantissa bits corresponding to our conversion target. Consequently, subtracting the FP16 value with exponent 50 and mantissa 0, will give us the floating point representation of exactly our 4 target bits, unsigned. To make this value signed, we further have to subtract $8$, which we can however fuse directly into the last 3 bits of the total value we subtract. A similar strategy works for different bit positions.

Modern GPUs can simultaneously compute with two separate 16-bit operands packed into a single 32-bit register. Hence, we can efficiently dequantize two INT4s in an INT32 at the same time, using the just described procedure. Finally, we want to dequantize directly into the right register layout for subsequent Tensor Core calls. To do this, we again take advantage of the fact that $\mathbf{B}$ can be preprocessed offline and reorganize weights such that the 16-byte vector read by each thread contains precisely its necessary 8 quantized weights of 4 separate $16\times16$ Tensor Core blocks. Additionally, within an INT32, weights are stored interleaved, according to the pattern 64207531, to power the previously mentioned parallel decoding.


At the innermost level, we accumulate the results of an $M \times 16$ times $16 \times 64$ matmul. We execute this accumulation column-wise, emitting $16\times16$ times $16\times8$ Tensor Core \code{mma.sync} instructions. This has the advantage over row-wise execution that we can pipeline the dequantization of the next $\mathbf{B}$ operand with the Tensor Core math of the current column.

\paragraph{Groups and Instruction Ordering.}
For per-output quantization, we can simply scale the final output once before the global write-out. An interesting observation is that despite these loads not being asynchronous to any computation, it is still critical to perform them via \code{cp.async} followed by an immediate \code{wait\_group} instruction, to avoid unfavorable main loop instruction reordering by the compiler.

With grouped quantization, which is crucial to maintain the good accuracy, we have to load and apply scaling during the main loop. First, we reorganize scale storage in a similar way as quantized weights (see above), such that the scales required by the same type of thread, for different $16 \times 16$ blocks, are packed together and can be loaded from shared memory as a single 16-byte vector. In principle, for group-size 128 and a $\mathbf{B}_\text{sm}$ tile shape of $64 \times 256$, we only need to global and shared memory load new scales \emph{once every other tile} (and here only once during the first sub-tile). However, it appears that the compiler is rather brittle to such irregularities in perhaps the most critical section of the code, leading to unfavorable instructions orderings with $10 - 20\%$ overall slow-down in some shape settings. Instead, we find that reloading scales from shared memory for \emph{every sub-tile} maintains peak performance. Doing this adds some technically unnecessary shared memory loads, but there is sufficient extra bandwidth to support this at no overhead, while it otherwise preserves the compiler's well pipelined instruction ordering for non-grouped quantization.

\paragraph{Striped Partitioning.}
With all the techniques described so far, we can reach near optimal compute and bandwidth performance, provided matrices are large and can be \emph{perfectly partitioned} across all SMs over the $N$ axis. In practice, this is rarely the case. The standard remedy in such a situation is to also partition across the $K$ dimension, but for many popular layer shapes and GPU combinations we would need a lot of additional splits to reach an even distribution without significant wave quantization. This in turn adds many global reduction steps, with additional overhead.

\begin{figure}[h!]
    \centering
    \includegraphics[width=.3\textwidth]{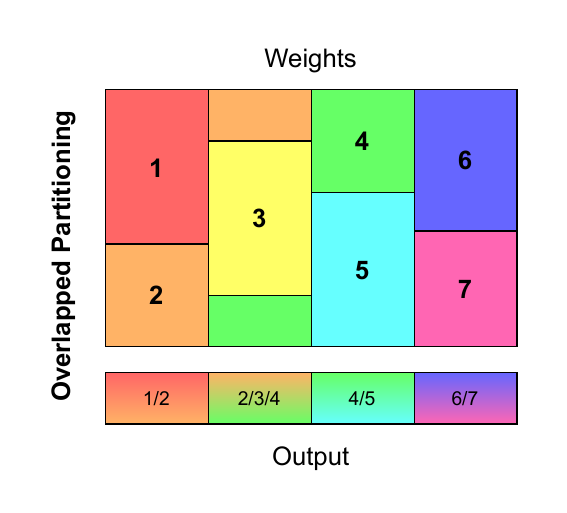}
    \caption{MARLIN's striped partitioning scheme.}
    \label{fig:marlin-partition}
\end{figure}

Instead, we opt for a \emph{striped} partitioning scheme where the ``stripes'' of $\mathbf{B}$ processed by an SM may span across multiple $\mathbf{C}_\text{sm}$ tiles (see also Figure~\ref{fig:marlin-partition}). Concretely, we first determine the number of $\mathbf{B}_\text{sm}$ tiles to be processed by each SM $T = \lceil \text{\#tiles} / \text{\#SMs} \rceil$ and then assign (up to) $T$ tiles column-wise starting top-left. Crucially, if we reach the bottom of a tile column but the current SM does not yet own $T$ tiles, we proceed by assigning tiles from the top of the next tile column; in other words, stripes can span across multiple columns. This ensures a roughly uniform distribution of tiles across all SMs, while minimizing the number of required global reduction steps. This strategy is similar to stream-k partitioning \cite{osama2023stream}.

We implement the global reduction between stripes of the same tile column serially, from bottom to top. The latter approach is most efficient since the bottom-most SM will have its results fastest and the top-most slowest in the presence of any column spill-over. We perform the reduction in FP16 directly in the output buffer to maximize L2 cache hits and thus minimize any global read overheads. This also keeps the operation essentially in-place, requiring only a small extra lock buffer for synchronization.

Finally, we note that for batchsizes $\gg 64$, we can virtually replicate $\mathbf{B}$ for the striped index calculations, followed by a modulo operation to move back into the original matrix, and advance the $\mathbf{A}$ pointer to the corresponding size-64 input batch segment. This results in significantly less global reductions for large input batchsizes (as occur during LLM prefills) and improves compute throughput in this setting.

\subsection{GPTQ Modifications}

The quantization format used by MARLIN, designed for peak inference efficiency, is slightly different than the original GPTQ implementation~\cite{frantar2022gptq}, yet still produces highly accurate models. We also integrate two small improvements into GPTQ: (a) picking group scales by searching for optimal group-wise clipping thresholds similar to \cite{lin2023awq}, and (b) supporting calibration sequences of variable length. We have found these modifications to yield small but consistent accuracy improvements over standard GPTQ, while having the advantage of higher performance. (We also provide a simple conversion script between model formats.) 

\begin{figure}[h!]
    \centering
    \includegraphics[width=0.85\linewidth]{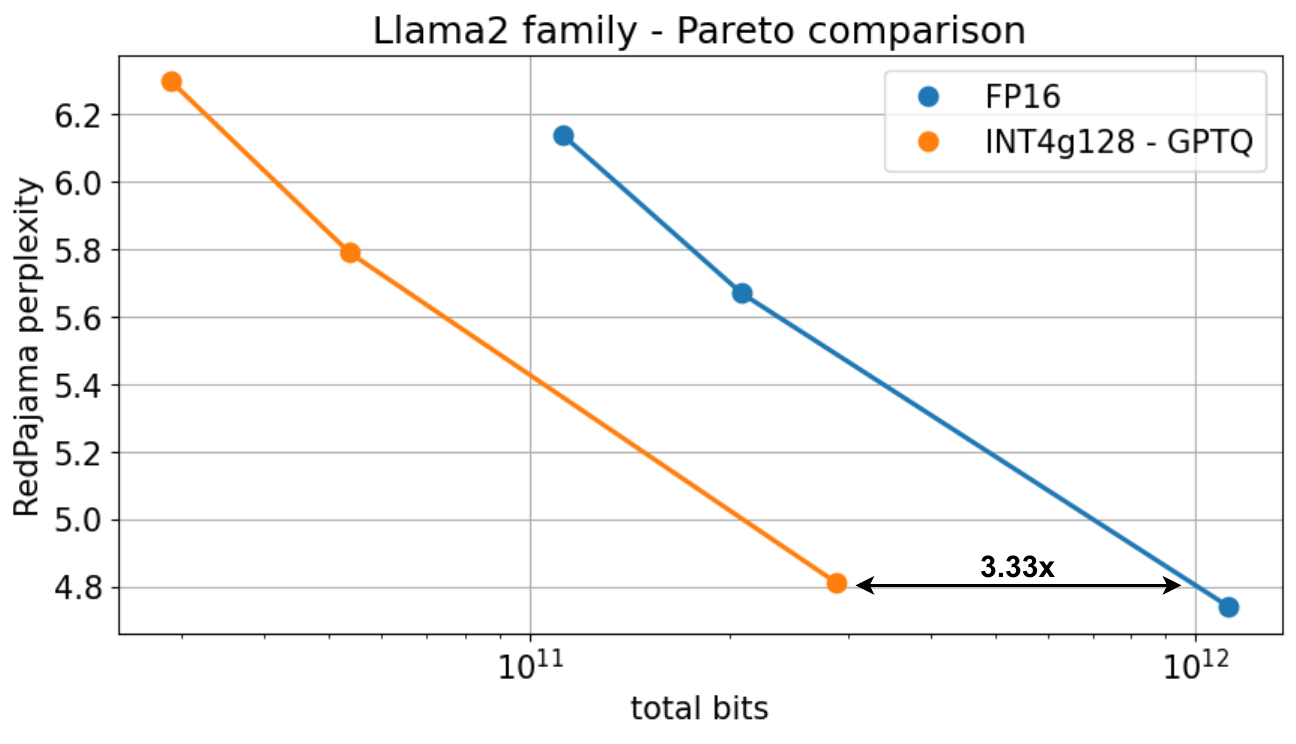}
    \caption{Pareto curve of Llama2 models quantized to the MARLIN format via GPTQ.}
    \label{fig:gptq-marlin}
\end{figure}

Figure~\ref{fig:gptq-marlin} illustrates perplexity (lower is better) versus model size in bits for our variant of GPTQ versus the original uncompressed models. This shows that MARLIN-quantized models are $\approx 3.33 \times$ smaller at the same perplexity as the uncompressed baseline. While this is not lossless (the ideal gain would be $3.87\times$ at this bit-width and group size), it is a significant improvement, especially given MARLIN's high inference efficiency.

\section{The Sparse-MARLIN Kernel}

To further improve FLOPS/Byte ratios, we can integrate a $2$:$4$ sparsity scheme on top of the  4-bit quantized weight representation. 
For background, the Sparse Tensor Cores (SPTCs) in the NVIDIA Ampere architecture provide an effective means to execute $50\%$ sparse matrices on specialized hardware units designed for sparse computation. 
Yet, to harness SPTCs, certain modifications and extensions to the previously described MARLIN kernel are required. 

First, to accommodate the constraints of the~\texttt{mma.sp} instruction, which enables the utilization of the SPTCs and requires sparse matrices as the Left-Hand-Side (LHS) operand in the tensor operation~\cite{nvidia-mma-sp}, we have designed new specific data layouts. 
Specifically, the problem of multiplying $\mathbf{A}$ with $\mathbf{B}$ is now reformulated under-the-hood as solving $\left( \mathbf{B}^{\top} \mathbf{A}^{\top} \right)^{\top}$ to produce $\mathbf{C}$.\footnote{Continuing with notation in Section~\ref{sec:marlin-kernel}, this is multiplying an $N\times K$ matrix (weight) with an $K\times M$ matrix (activation) to produce an $M\times N$ output.}
However, this reformulation retains all the techniques and optimizations from the dense MARLIN kernel design previously described.
Note that $\mathbf{B}$ can be preprocessed offline as needed, while $\mathbf{A}$ can be transposed on-the-fly in SMEM with native support via the \code{ldmatrix} instruction and its \code{.trans} optional qualifier, without incurring performance degradation.

Next, we describe the two new data structures necessary for encoding $2$:$4$ sparse matrices in Sparse-MARLIN, along with their adaptations tailored to address this particular problem: 
(1) the non-zero values structure, and 
(2) the metadata indices structure.

\begin{figure}[ht]
    \centerline{\includegraphics[width=\linewidth]{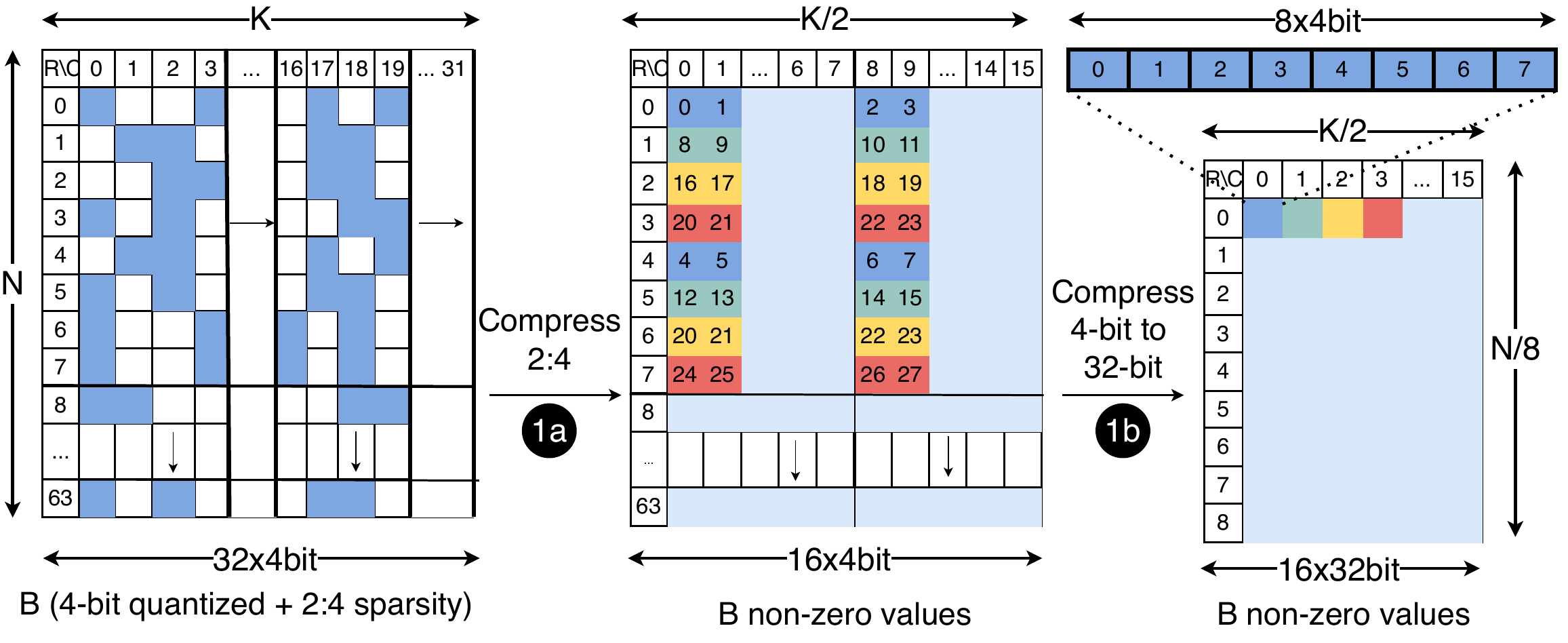}}
    \caption{Sparse-MARLIN non-zero values structure layout}
    \label{fig:2_4_layout}
\end{figure}

\paragraph{Quantized non-zero values.}
Figure~\ref{fig:2_4_layout}, left side, illustrates a 4-bit quantized matrix $\mathbf{B}$ of size $N\times K$ which has been pruned to $2$:$4$ sparsity. 
The compressed representation of this matrix, depicted in~\encircles{$1_{a}$}, will have half the size of the original one in the inner dimension, that is, $N\times K/2$.
However, as each value is a $4$-bit element, we can apply the dense MARLIN compression approach on top of this to further compress $8$ elements in a $32$-bit value, as depicted in~\encircles{$1_{b}$}, with a final size of $N/8\times K/2$.

To maximize memory efficiency, since the weights remain static during inference, each $64\times 16$ tile is reshuffled so that each thread loads and stores elements in contiguous memory positions, similar to dense MARLIN.
Continuing with Figure~\ref{fig:2_4_layout},
the colored elements in the non-zero values structure represent an example of the elements supplied by thread $T_0$ for one $64\times 16$ block of $\mathbf{B}$. 
The paired colors denote elements processed within the same~\texttt{mma.sp} instruction, necessitating $4$ iterations to compute all elements.
This layout ensures the widest 128-bit instructions ($4$ consecutive $32$-bit elements per thread, as shown in~\encircles{$1_{b}$}) when loading the non-zero value structure from GMEM. 

Furthermore, due to the redefinition of the product and since the output $\mathbf{C}$ will be an FP16 matrix of shape $M\times N$, this layout also ensures $128$-bit instructions (e.g., first eight consecutive output elements in column $0$ stored in $T_0$ registers) when storing the results transposed from RF to GMEM.
Thus, this reformulation of the product not only stores the results transposed without incurring performance degradation, but further improves the efficiency of output writing compared to the baseline dense design.

\begin{figure}[t!]
    \centerline{\includegraphics[width=\linewidth]{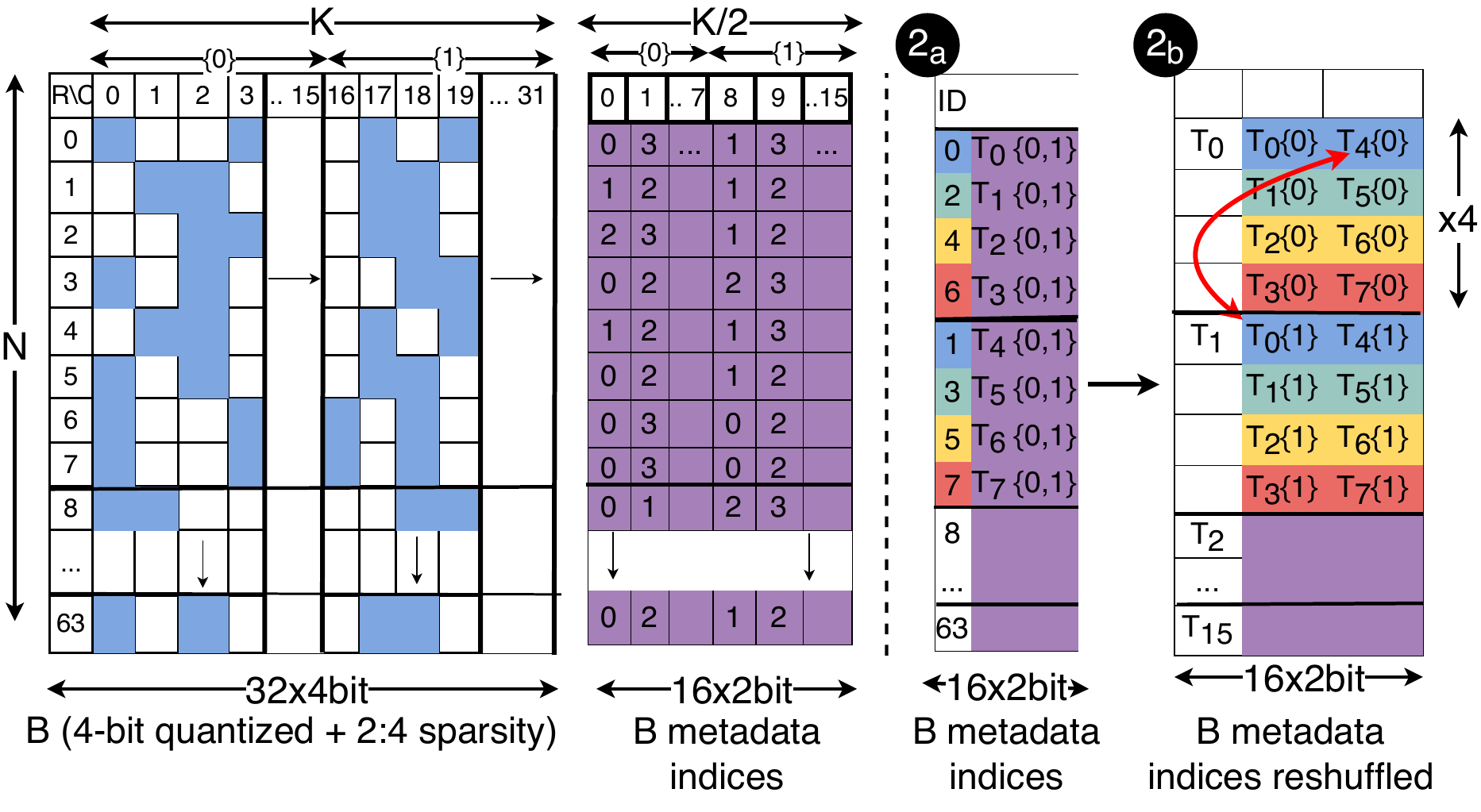}}
    \caption{Sparse-MARLIN metadata indices layout.}
    \label{fig:2_4_layout_meta}
\end{figure}

\paragraph{Metadata indices.}
In order to select the elements from the Right-Hand-Side (RHS) operand $\mathbf{A}$ that will be necessary in the sparse computation, a metadata structure containing the indexes of non-zero elements in the original matrix is required.
Figure~\ref{fig:2_4_layout_meta}, left side, illustrates the metadata indices structure of $\mathbf{B}$. 
Since this is $2$:$4$ sparsity, indices will be in the range $0\sim 3$, encoded with $2$ bits. 
Based on the data layout described in Figure~\ref{fig:2_4_layout}, and considering the sparsity selector constraints of the~\texttt{mma.sp} instruction, we propose a new data layout for the metadata structure.
The sparsity selector indicates the thread(s) within a group of four consecutive threads that will contribute the metadata for the entire group. In the example depicted in Figure~\ref{fig:2_4_layout_meta}, the sparsity selector can be either $0$ (threads $T_0$, $T_1$) or $1$ (threads $T_2$, $T_3$).

First, we have to reorder the rows based on the non-zero values structure previously described, as shown in~\encircles{$2_{a}$}. This allows us to use 128-bit load instructions from GMEM to SMEM. Then, in order to load the data from SMEM to RF bank-conflict free, we perform a single~\texttt{ldmatrix} instruction which will contain all the information for the next four~\texttt{mma.sp} operations to be executed, and which will distribute the information across threads $T_0\sim T_3$ as required. However, a previous data reshuffling is needed, as~\encircles{$2_{b}$} shows. This way, threads $T_0$, $T_1$ will contain the information for the first two iterations, and threads $T_2$, $T_3$ will have the information for the two remaining ones. Remark that all this pre-processing is done offline once, without runtime overhead.

\section{Experimental Results}

\subsection{Kernel Benchmarks}
\label{sec:marlin-benchmarks}

In our first set of experiments, we examine the efficiency of MARLIN relative to an ideal kernel, and compare its performance with other popular 4-bit inference kernels, notably the well-optimized PyTorch kernel~\cite{paszke2019pytorch}, the AWQ kernel~\cite{lin2023awq}, the open-source ExLlamaV2 kernel~\cite{exllamav2, chavan2024faster}, and the bits-and-bytes kernel~\cite{dettmers2022llm}, 
on a large matrix that can be ideally partitioned on a target GPU. For this, we choose the NVIDIA A10 GPU, which is popular for inference workloads. This allows all kernels to reach close to their best possible performance. All kernels are executed at 4-bit and groupsize 128. (However, scale formats are not 100\% identical, due to small differences between the methods.) 

Figure~\ref{fig:marlin-peak} shows our results for a large 72k $\times$ 18k matrix.
We observe that, while existing kernels achieve relatively close to the optimal $3.87\times$ speedup at batchsize 1 (note the 0.125 bits storage overhead of the group scales), their performance degrades quickly as the number of inputs is increased. In contrast, MARLIN delivers close to ideal speedups at all batchsizes, enabling the maximum possible $3.87\times$ speedup up to batchsizes around 16-32, and tailing off as we transition from the memory- to the compute-bound matmul regime.

\begin{figure}[h!]
    \centering
    \includegraphics[width=0.9\linewidth]{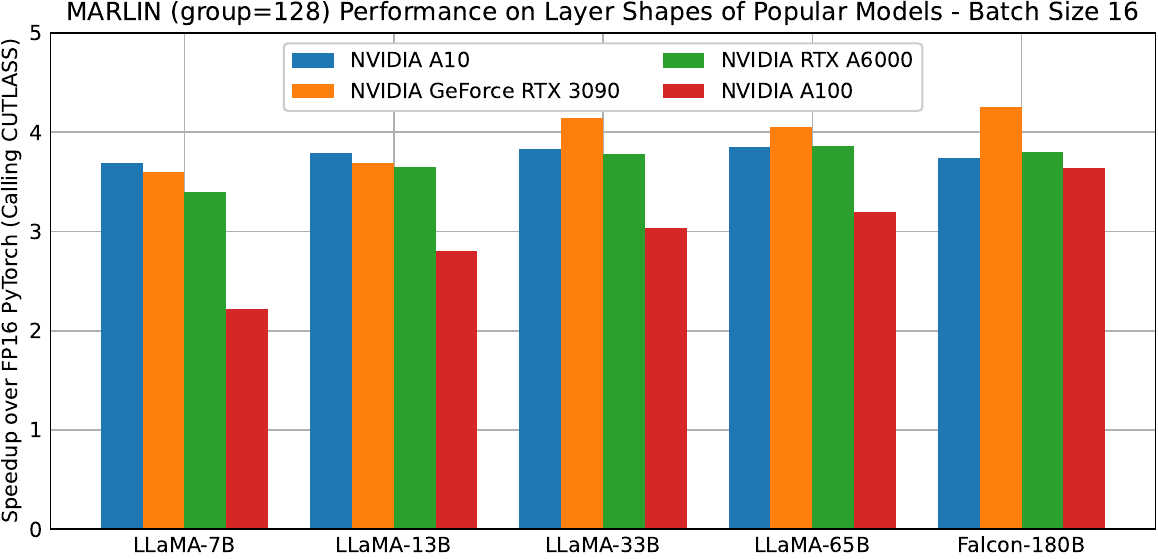}
    \caption{MARLIN performance across real layer shapes of popular models.}
    \label{fig:marlin-models}
\end{figure}

Due to its striped partitioning scheme, MARLIN brings strong performance also on real (smaller) matrices and various GPUs. This is demonstrated in Figure~\ref{fig:marlin-models}, where we benchmark, at batchsize 16, the overall speedup (relative to FP16) across some individual linear layers in popular open-source models~\cite{touvron2023llama2, falcon2023}, showing similar trends. We observe better speedups on commodity GPUs such as the NVIDIA GeForce RTX 3090, and lower speedups on the flagship NVIDIA A100 GPU; this is because the latter has significantly higher GMEM bandwidth and compute, which makes overheads such as pipeline startup latency or suboptimal partitioning relatively bigger, especially on small matrices.

\begin{figure}[h!]
    \centering
    \includegraphics[width=0.9\linewidth]{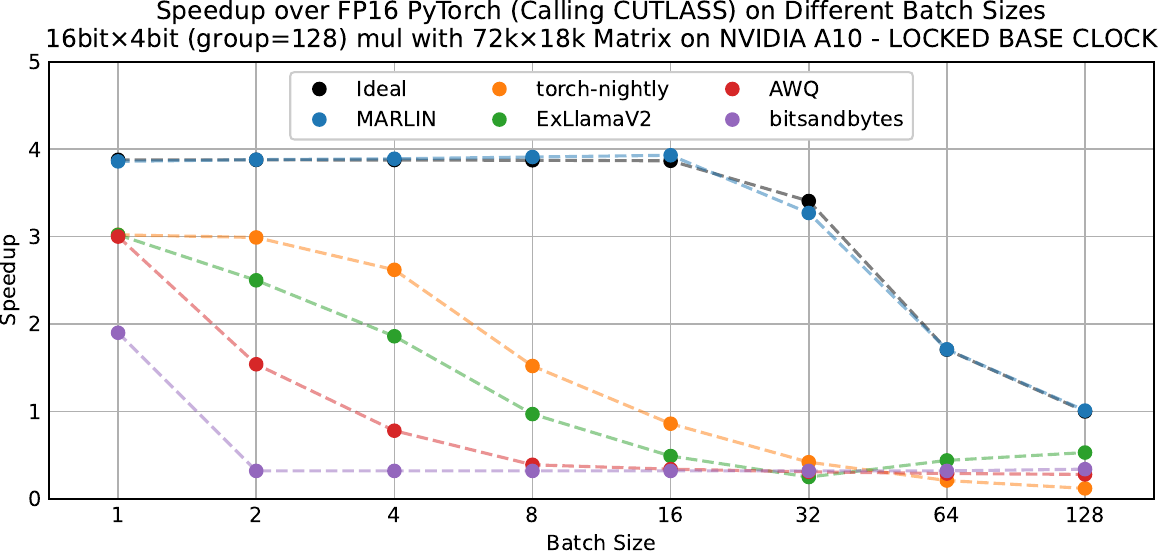}
    \caption{Sustained performance of MARLIN compared with other popular open-source kernels.}
    \label{fig:marlin-sustained}
\end{figure}

Next, we also study what performance can be sustained over longer periods of time, at locked base GPU clock, as this is a probable scenario in a production setting.  Interestingly, we find that reduced clock speeds significantly harm the relative speedups of prior kernels, but have no effect on MARLIN's virtually optimal performance (relative to the lower clock setting). This can be observed in Figure~\ref{fig:marlin-sustained}.

Finally, we also tested how MARLIN performed on \emph{very large batch sizes}, corresponding to the initial prompt-processing ``prefill'' inference step while running on a powerful GPU like the A100. We observed that, even in this case, MARLIN is nearly identical to an uncompressed compute-bound matmul up to batch size 1024, with only $\approx 10\%$ slow-down at even larger input shapes. We leave optimizations in this particular scenario for future work. 



\begin{figure}[t]
    \centering
    \includegraphics[width=0.9\linewidth]{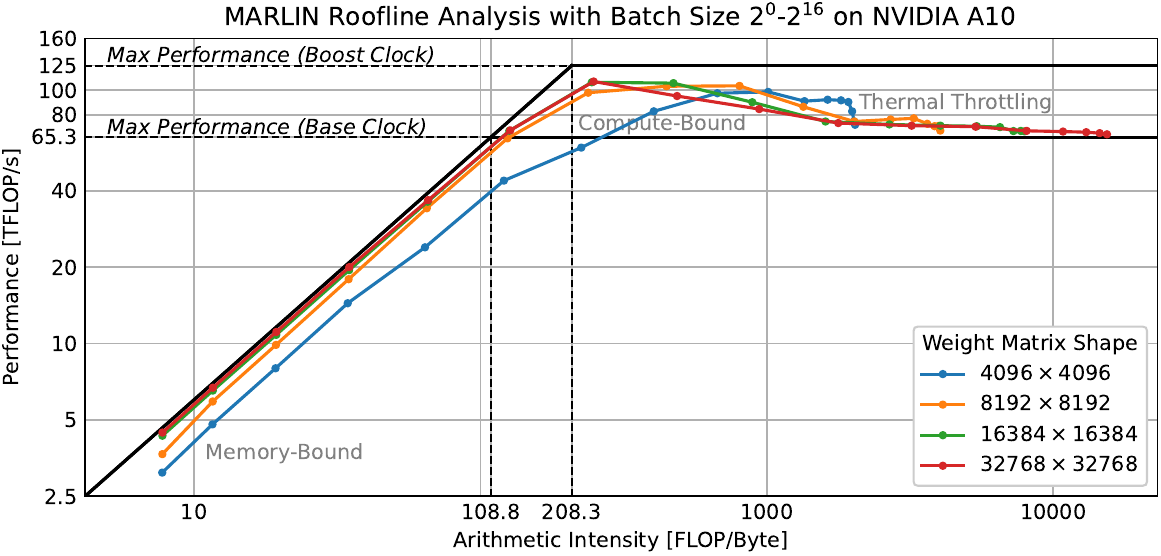}
    \caption{Roofline analysis for the MARLIN kernel, across four different matrix shapes.}
    \label{fig:marlin_roofline}
\end{figure}

\label{sec:roofline}
\paragraph{Roofline Analysis.}
To gain a deeper understanding of the computational efficiency of MARLIN, we perform a roofline analysis, which is a widely accepted methodology for evaluating performance potential. Figure~\ref{fig:marlin_roofline} shows the roofline analysis of the matrix multiplication operation performed on the MARLIN kernel on an NVIDIA A10 GPU. Several typical weight matrix sizes ($2^{12}, 2^{13}, 2^{14}, 2^{15}$) are tested during the analysis. The markers on curves are the profiling results with different input batch sizes ($2^{0}, 2^{1}, ..., 2^{16}$). 

First, note that the GPU itself offers different performance levels, depending on whether the boost clock is enabled and \emph{can be sustained} (see horizontal lines). 
Generally, we first observe that batchsizes smaller than 64 are memory-bound, while the larger batchsizes are compute-bound, confirming our prior intuition. Further, the MARLIN kernel achieves strong hardware utilization across matrix sizes and arithmetic intensities, with the best results for larger matrices. Interestingly, we observe that for time intensive computations (large matrices and batchsizes), the GPU's clock rate is automatically throttled and FLOP/s correspondingly drop towards the base clock limit.

\begin{figure}[ht]
    \centering
    \includegraphics[width=0.9\linewidth]{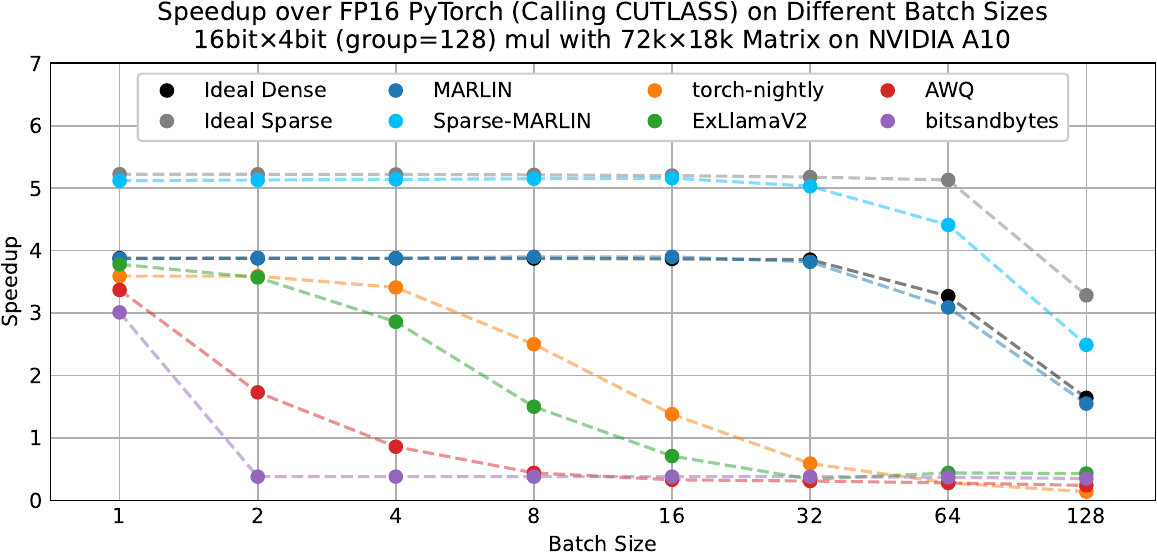}
    \caption{Peak performance of MARLIN and Sparse-MARLIN versus other popular open-source kernels.}
    \label{fig:smarlin-peak}
\end{figure}

\begin{figure}[ht]
\centering
\includegraphics[width=0.9\linewidth]{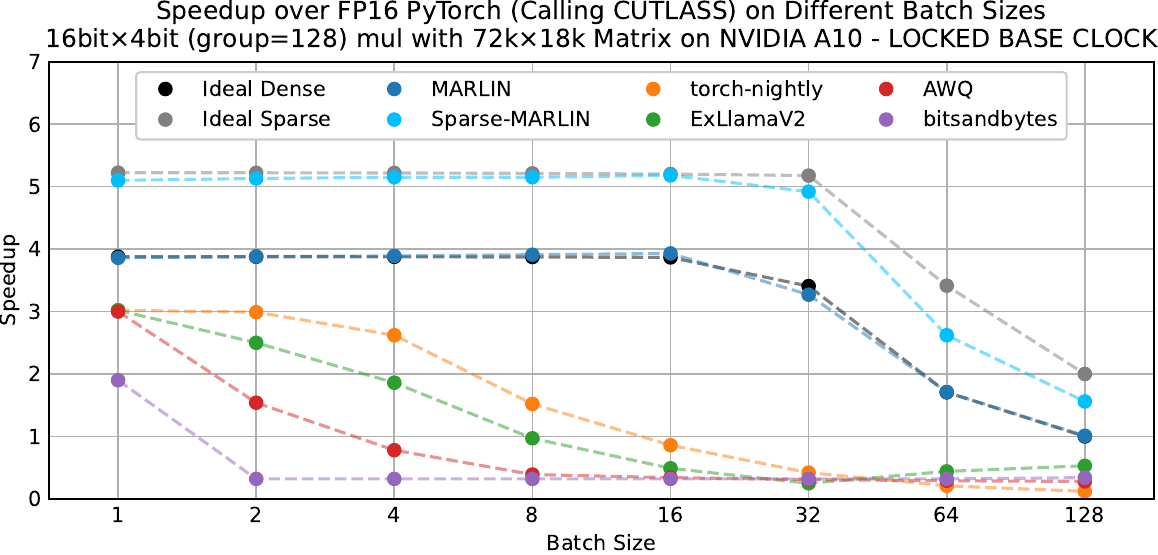}
\caption{Sustained performance of MARLIN and Sparse-MARLIN compared with other popular open-source kernels.}
\label{fig:smarlin-sustained}
\end{figure}

\paragraph{Performance of Sparse-MARLIN.}
We now examine improvements due to 2:4 sparsity. 
Figure~\ref{fig:smarlin-peak} shows peak performance of Sparse-MARLIN compared to ideal lines, dense variants, and popular open-source kernels, while Figure~\ref{fig:smarlin-sustained} shows sustained performance. These  figures again demonstrate strong performance of this implementation, thus validating the extensibility of our design to other formats.

\subsection{End-to-End Experiments}

Next, we validate our approach end-to-end (i.e., on full models) in a realistic LLM serving setting. For this, we examine the performance of MARLIN and its sparse variant when integrated into the popular open-source vLLM serving engine~\cite{vllm}. 


\paragraph{Accuracy.} In Table~\ref{tab:llama-2-7b_accuracy_benchmarks} we briefly examine the accuracy difference between the baseline and sparse and sparse-quantized versions of Llama2. While recovering model accuracy is not our focus in this paper, we note that these results show that accuracy can be well-recovered under compression.

\begin{table}[ht]
\centering
\small
\caption{Llama-2-7B accuracy for the original model and quantized/sparse models. The INT4 model is used for MARLIN, generated by our version of GPTQ~\cite{frantar2022gptq}. The INT4 + 2:4 model is used for Sparse-MARLIN, generated by SparseGPT~\cite{frantar2023sparsegpt}; it has higher metrics than the original because it is further fine-tuned on synthetic data, via Knowledge Distillation.}
\label{tab:llama-2-7b_accuracy_benchmarks}
\centering
\begin{tabular}{ c | c | c c c }
\hline
\textbf{Benchmark} & \textbf{Metric} & \textbf{Baseline} & \textbf{INT4} & \textbf{INT4 + 2:4} \\
\hline
MMLU & 5-shot & 47.88 & 43.59 & 48.81 \\
WinoGrande & 5-shot & 71.82 & 68.75 & 73.09 \\
ARC-Challenge & 25-shot & 51.19 & 48.55 & 53.67 \\
\hline
\multicolumn{2}{c|}{\textit{Mean}} & \textit{56.96} & \textit{53.63} & \textit{58.52} \\
\hline
\end{tabular}
\end{table}

\paragraph{Integration with vLLM.} 
We first compare the end-to-end performance of MARLIN and Sparse-MARLIN with the default 16-bit kernel via a vLLM integration. The GPTQ-quantized models are used for MARLIN and Sparse-MARLIN, while the original unquantized models are used for the baseline. We perform the benchmark using 64 input tokens per sequence in a batch and let the model generate another 64 tokens for each sequence. We intentionally make the sequence length small so that it can fit into the memories of more types of GPUs and the computation cost is not dominated by attention calculations.

\begin{figure}[t]
    \centering
    \includegraphics[width=0.9\linewidth]{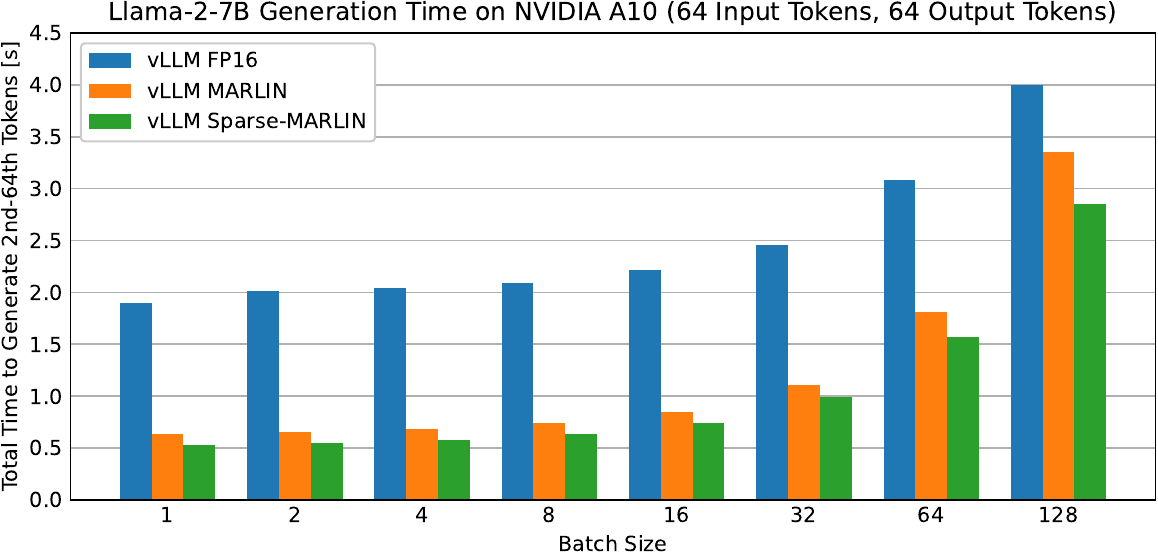}
    \caption{End-to-end generation time of MARLIN and Sparse-MARLIN compared with the vLLM FP16 baseline.}
    \label{fig:e2e_generation_time_llama-2-7b}
\end{figure}

Figure~\ref{fig:e2e_generation_time_llama-2-7b} shows the total time it takes for the Llama-2-7B model to generate new tokens on NVIDIA A10 in the generation phase, which reflects the output token throughput. The MARLIN kernel has a speedup up to approximately $3\times$, while Sparse-MARLIN provides an additional $1.2\times$ end-to-end speedup on top of MARLIN. The reduction in speedup relative to our prior per-layer experiments is due to the various additional overheads of inference, outside the linear layers that we accelerate.

\paragraph{GPU and Model Types.} Next, Table~\ref{tab:e2e_generation_speedup} shows MARLIN speedups under a variety of settings using several popular (quantized) models on different GPU types. In addition, for large models, we also examine the impact of sharding the weight matrices across multiple GPUs, supported by vLLM. 

{
\newcommand{\tabhead}[1]{\textbf{#1}}
\begin{table*}[ht]
\centering
\caption{End-to-end generative speedup of MARLIN compared to vLLM's FP16 baseline.}
\label{tab:e2e_generation_speedup}
\centering
{
\begin{tabular}{ c | c | c | c c c c c c c c }
\hline
\multirow{2}{*}{\tabhead{LLM Model}} & \multicolumn{2}{c|}{\tabhead{GPU}} & \multicolumn{8}{c}{\tabhead{Batch Size}} \\
\cline{2-11}
& \tabhead{Type} & \tabhead{\#} & \tabhead{1} & \tabhead{2} & \tabhead{4} & \tabhead{8} & \tabhead{16} & \tabhead{32} & \tabhead{64} & \tabhead{128} \\
\hline
\multirow{2}{*}{Llama-2-7B} & \multirow{1}{*}{A10} & 1 & 2.93 & 3.19 & 3.02 & 2.90 & 2.74 & 2.26 & 1.78 & 1.20 \\
\cline{2-11}
& \multirow{1}{*}{3090} & 1 & 2.69 & 2.58 & 2.52 & 2.43 & 2.30 & 1.84 & 1.28 & 1.11 \\
 \hline
\multirow{1}{*}{Llama-2-13B} & \multirow{1}{*}{A6000} & 1 & 3.17 & 3.13 & 3.07 & 2.97 & 2.77 & 2.39 & 2.01 & 1.23 \\
\hline
\multirow{1}{*}{Yi-34B} & \multirow{1}{*}{A100} & 1 & 2.90 & 2.88 & 2.86 & 2.79 & 2.69 & 2.36 & 1.71 & 1.18 \\
\hline
\multirow{5}{*}{Llama-2-70B} & \multirow{2}{*}{A6000} & 4 & 2.84 & 2.87 & 2.93 & 2.84 & 2.67 & 2.02 & 1.74 & 1.23 \\
& & 8 & 2.10 & 2.04 & 2.26 & 2.19 & 2.06 & 1.43 & 1.42 & 1.11 \\
\cline{2-11}
& \multirow{3}{*}{A100} & 2 & 2.55 & 2.59 & 2.57 & 2.53 & 2.42 & 2.08 & 1.52 & 1.19 \\
& & 4 & 2.02 & 1.97 & 2.01 & 1.99 & 1.97 & 1.63 & 1.29 & 1.14 \\
& & 8 & 1.38 & 1.42 & 1.44 & 1.44 & 1.44 & 1.19 & 1.04 & 1.07 \\
\hline
\multirow{2}{*}{Falcon-180B} & \multirow{1}{*}{A6000} & 8 & 2.24 & 2.06 & 1.90 & 1.67 & 1.55 & 1.37 & 1.38 & 1.27 \\
\cline{2-11}
& \multirow{1}{*}{A100} & 8 & 1.76 & 1.74 & 1.76 & 1.75 & 1.70 & 1.65 & 1.29 & 1.08 \\
\hline
\end{tabular}
}
\end{table*}
}

We find that MARLIN improves the performance in all scenarios. The largest speedups happen when inference is memory-bound (up to batchsize $\approx 16$) and the GPUs are weaker or fewer in number. The finding is natural as overheads are relatively more costly when the absolute runtime of core operations is lower.
Thus, MARLIN is particularly beneficial in resource-constrained settings.

\begin{figure}[ht]
    \centering
    \includegraphics[width=0.9\linewidth]{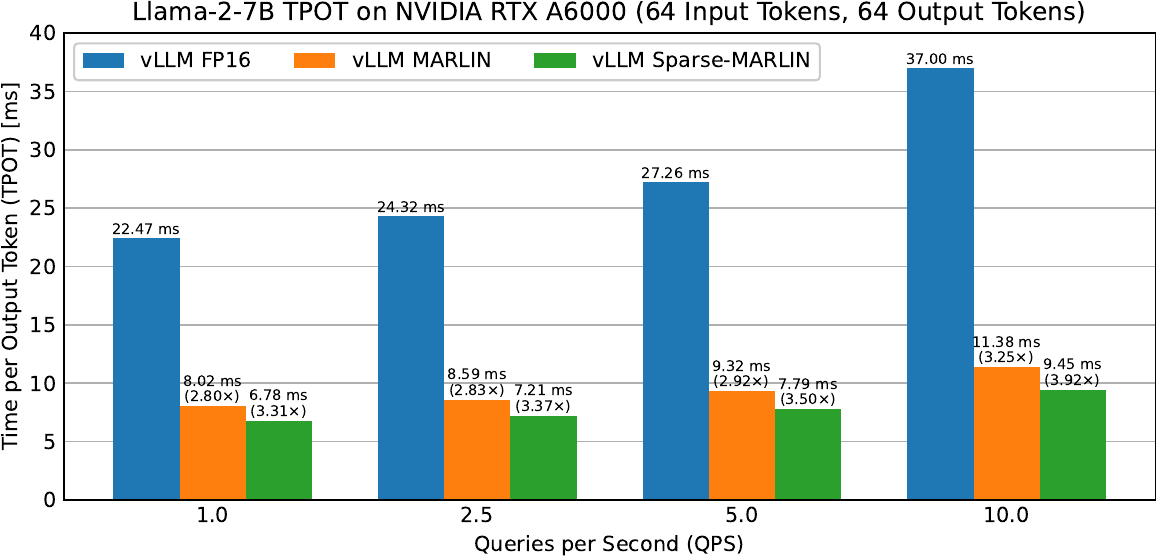}
    \caption{Serving benchmark (TPOT, Time Per Output Token) of MARLIN and Sparse-MARLIN compared with the vLLM FP16 baseline.}
    \label{fig:tpot_llama-2-7b}
\end{figure}

\paragraph{Client Counts.} Finally, we perform a serving benchmark in a simulated server-client setting and measured the standard TPOT metric (Time Per Output Token, the average latency to generate an output token for each queried sequence) under different querying intensities (queries-per-second or QPS), which influences the average batch size per inference. 

\begin{figure}[ht]
\centering
\includegraphics[width=0.9\linewidth]{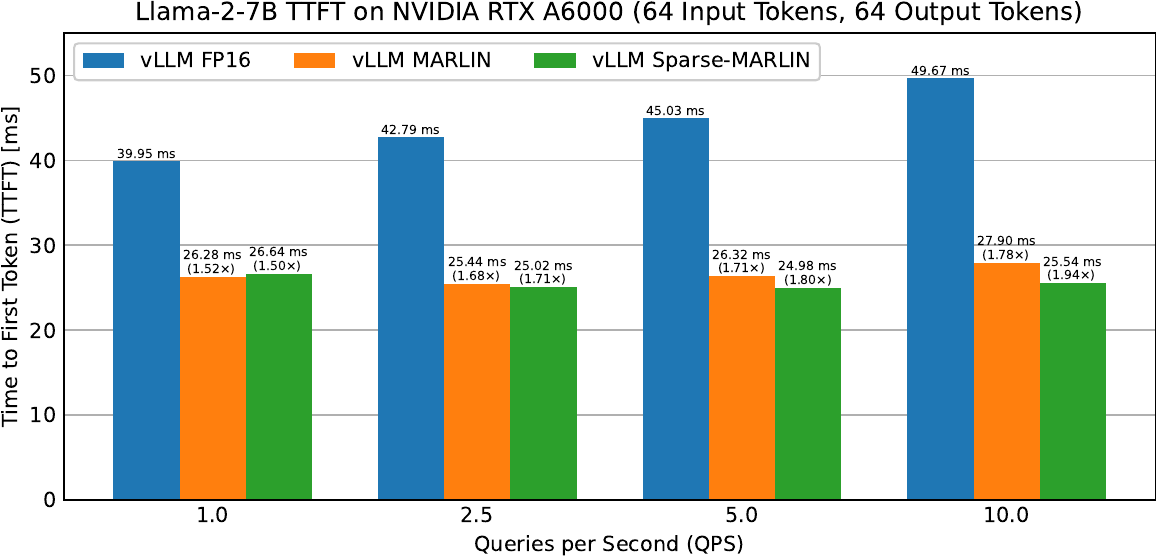}
\caption{Serving benchmark (TTFT, Time To First Token) of MARLIN and Sparse-MARLIN compared with the vLLM FP16 baseline.}
\label{fig:ttft_llama-2-7b}
\end{figure}

Figure~\ref{fig:tpot_llama-2-7b} shows the results of Llama-2-7B on an NVIDIA RTX A6000 GPU. We observe that MARLIN achieves approximately $2.8\times$ latency reduction, while Sparse-MARLIN provides about $3.3\times$ speedup, noticing that speedups relative to FP16 are stable across QPS values. 
Observe that the speedup relative to the baseline \emph{increases} as we increase number of queries per second (QPS). 
We believe that the explanation for this phenomenon is the following: due to reduced memory movement, MARLIN allows for lower average per-query latency; in turn, this implies that the \emph{average batch size} at which MARLIN executes is \emph{lower} than the average batch size for FP16. In turn, this means that the relative gains of MARLIN will increase as we scale up the number of clients. 
Figure~\ref{fig:ttft_llama-2-7b} shows that MARLIN can also lead to improvements in the case where prompt processing is also taken into account, where we measure time to first token (TTFT).

\section{Related Work}

Due to space constraints, we focus on closely related work about providing efficient support for quantized LLM inference. As noted previously, there is already significant work on accurate LLM weight quantization, with popular methods such as GPTQ~\cite{frantar2022gptq} and AWQ~\cite{lin2023awq}, as well as explorations of round-to-nearest (RTN) quantization~\cite{dettmers2022case}, which is usually less accurate. 
The MARLIN parallelization approach can be generalized to these quantization methods. In fact, since the original release of our kernel for the GPTQ format, a version of MARLIN supporting AWQ has been introduced independently in vLLM~\cite{vllm-awq-marlin}.  

More broadly, LLM quantization methods can also consider compressing both weights and activations~\cite{dettmers2022llm}, with advanced methods such as SmoothQuant~\cite{xiao2022smoothquant} or QuaRot~\cite{ashkboos2024quarot}. However, quantization of activations tends to be more complex, due to the emergence of large ``outlier'' values~\cite{dettmers2022llm}. 
As such, those approaches tend to either target lower 8bit precision, or require more complex additional processing steps, such as via rotation matrices~\cite{ashkboos2024quarot}. 
The MARLIN approach is extensible to this case, for instance, recent independent follow-up to MARLIN extended our approach to the case where activations are quantized to 8 bits, while weights are quantized to 4 bits~\cite{zhang2024qqq}.

\section{Discussion and Future Work} 

We have presented MARLIN, a general approach for implementing  mixed-precision kernels for LLM generative inference, which achieves near-optimal efficiency by leveraging new GPU hardware instructions and parallelization techniques. 
Specifically, we have shown that MARLIN and its sparse counterpart reach near-optimal per-layer efficiency, and can lead to speedups of up to $3\times$ in real-world deployment scenarios, at moderate accuracy impact. 
In terms of future work, a natural direction is  investigating MARLIN support for the recently proposed and more complex techniques for ``extreme'' compression via vector quantization~\cite{chee2023quip, egiazarian2024extreme}, which require more complex decompression. Another direction is to investigate MARLIN support for additional forms of mixed precision, such as the ones arising from activation compression or sparsity.

\section*{Acknowledgments}

The authors would like to thank the Neural Magic team, in particular Michael Goin, Alexander Matveev, and Rob Shaw, for support during the writing of this paper, in particular with the vLLM integration. This research was supported in part by generous grants from NVIDIA and Google.


\bibliography{references}
\bibliographystyle{arxiv}



\end{document}